\documentclass[conference]{IEEEtran}
\IEEEoverridecommandlockouts
\usepackage{cite}
\usepackage{amsmath,amssymb,amsfonts}
\usepackage{algorithmic}
\usepackage{algorithm2e}
\usepackage{graphicx}
\usepackage{caption}
\usepackage{subcaption}
\usepackage{textcomp}
\usepackage{tabularx}
\usepackage{multirow}
\usepackage[dvipsnames]{xcolor}
\newcommand*{\vic}[1]{\textcolor{black}{#1}}

\def\BibTeX{{\rm B\kern-.05em{\sc i\kern-.025em b}\kern-.08em
    T\kern-.1667em\lower.7ex\hbox{E}\kern-.125emX}}
\begin{document}

\newcommand{\tl}[1]{{\leavevmode\color{black}{#1}}}

\title{Keep It Simple: Fault Tolerance Evaluation of Federated Learning with Unreliable Clients
\thanks{Identify applicable funding agency here. If none, delete this.}
}

\author{\IEEEauthorblockN{Victoria Huang\IEEEauthorrefmark{1}, Shaleeza Sohail\IEEEauthorrefmark{2}, Michael Mayo\IEEEauthorrefmark{3}, Tania Lorido Botran\IEEEauthorrefmark{4}, Mark Rodrigues\IEEEauthorrefmark{3}, \\Chris Anderson\IEEEauthorrefmark{3}, Melanie Ooi\IEEEauthorrefmark{3}}
\IEEEauthorblockA{\IEEEauthorrefmark{1}\textit{National Institute of Water and Atmospheric Research}, Wellington, New Zealand \\
victoria.huang@niwa.co.nz}
\IEEEauthorblockA{\IEEEauthorrefmark{2}\textit{University of Newcastle}, New South Wales, Australia \\
shaleeza.sohail@newcastle.edu.au}
\IEEEauthorblockA{\IEEEauthorrefmark{3}\textit{University of Waikato},
Hamilton, New Zealand \\
Michael.mayo, Mark.rodrigues, Chris.anderson, Melanie.ooi@waikato.ac.nz}
\IEEEauthorblockA{\IEEEauthorrefmark{4}\textit{Roblox}, California, United States \\
tbotran@roblox.com}
}

\maketitle
% Back to the Drawing Board: A Critical Evaluation of Poisoning Attacks on Production Federated Learning

\begin{abstract}
Federated learning (FL), as an emerging artificial intelligence (AI) approach, enables decentralized model training across multiple devices without exposing their local training data. FL has been increasingly gaining popularity in both academia and industry. While research works have been proposed to improve the fault tolerance of FL, the real impact of unreliable devices (e.g., dropping out, misconfiguration, poor data quality) in real-world applications is not fully investigated. \tl{We carefully chose two representative, real-world classification problems with a limited numbers of clients to better analyze FL fault tolerance. Contrary to the intuition, simple FL algorithms can perform surprisingly well in the presence of unreliable clients.}

%Intuitively especially in scenarios with a limited number of clients}
%Experiments conducted using two real-world datasets show that even simple FL algorithms can perform . 
% in environments with unreliable power supplies, unstable communication, and even misconfigured clients. 
\end{abstract}


\section{Introduction}\label{sec:intro}

As an emerging learning paradigm, federated learning (FL) provides a new training method that allows data owners (also called \emph{clients}) to enjoy the performance of a jointly trained model without violating their data privacy. Whereas traditional machine learning (ML) follows a data-to-model approach (i.e. training data from various clients need to be centralized in one node), FL takes an opposite model-to-data approach. In FL, models trained by clients using their local data are shared to a centralized node (called \emph{server}), as opposed to the data. The server updates a global model by aggregating the received local models. The global model will be broadcasted to clients to replace their local models. This means that data never leaves the clients, thus preserving data privacy. Meanwhile, the local sites can take advantage of the improved accuracy resulting from the aggregated model trained across all of the data as opposed to local data only. FL therefore addresses data privacy and data sharing dilemma while potentially offering superior model performance. Given its promising features, FL has attracted much research attention~\cite{kang2020reliable}. 
% Different approaches have been proposed in the literature to improve communication efficiency [ref], enhance data protection [ref], and address data and client heterogeneity [ref].  

Although FL has great potential in real-world applications, there have been very few FL applications reported at the production level, with most of the work being at a proof-of-concept prototype level with synthetic datasets~\cite{lo2021systematic}. Meanwhile, implementing FL in a rural environment introduces multiple further complications,
all of which are currently active targets for research \cite{lim2020,patros2022rural,almurshed2022adaptive}. The complications can be broadly grouped into the following issues:
\begin{itemize}
    \item reducing communication overhead (e.g., using model compression~\cite{xu2022adaptive}).
    \item improving energy efficiency (e.g., modifying conventional FL algorithms to save 59\% energy cost~\cite{yang19}).
    \item heterogeneity of clients and where to place computation (e.g., some of the model aggregation can be performed at the edge servers to reduce the communication overhead and also the load on the cloud server~\cite{liu2020client}).
    \item \vic{impact of unreliable clients that drop in and out of the federation~\cite{mao2022safari,wang2022friends} or send low-quality data~\cite{tsouvalas2022federated,wang2022fednoil}.}
\end{itemize}
This last problem (unreliable clients) is largely unaddressed in the literature and posed as an open research challenge~\cite{lim2020}.

We therefore focus our study in this paper via a systematic evaluation on the robustness of FL to unreliable clients in rural applications. Note that in this paper, we consider different scenarios of unreliable clients related to \textit{infrastructure-level errors} and \textit{ML-specific inconsistencies}:
\begin{itemize}
    \item Infrastructure-level errors:
    \begin{itemize}
        % \item clients randomly and completely drop in and out of the federation. For example, IoT devices deployed in isolated areas with unstable power supplies.
        \item \vic{Clients randomly and completely drop in and out of the federation. This can be caused by unstable power supplies when clients are deployed in isolated areas. Clients may sometimes drop out during the training process and join back later.}
        % \item clients randomly and partially drop in and out. For example, a client may not be able to upload its locally trained model or receive the aggregated global model during the training due to an unreliable network connection.
        \item \vic{Clients randomly and partially drop in and out. Compared to completely dropping out, clients may sometimes fail to upload their locally trained models but they are still able to receive the aggregated global model or vice versa. This can happen when the network connection is unreliable.} 
    \end{itemize}
    \item ML-specific inconsistencies:
    \begin{itemize}
        \item Client misconfiguration. For example, the hyperparameters (e.g., learning rate) of one client are misconfigured and different from others.
        % \item Clients with low-quality data which can be mislabelled or low resolution. 
        \item \vic{Low-quality data. For example, the training data collected by the client may be low-quality due to hardware constraints or mislabelled due to human mistakes.}
    \end{itemize}
\end{itemize}

\tl{The ultimate impact of unreliable clients on a particular system will depend on other factors, such as the number of clients or the size of local dataset \cite{kamp2021federated}. There are efforts to establish metrics that quantify the influence of individual clients on the model outcome, but such metric fails to capture the overall reliability of FL algorithms\cite{xue2021toward}.}
\vic{Moreover, applications reported in the literature~\cite{nilsson2018performance,mcmahan17} usually involve hundreds of clients. It should be noted that in rural settings, there may only be a very limited number of clients involved, making the impact of unreliable clients potentially significant. Thus, we carefully select our target scenarios that involve \textit{a limited number of clients.}}
% In order to provide fine-grained results, the target scenarios must have \textit{a limited number of clients.}

In this paper, we develop a prototype system building on federated Function-as-a-Service using funcX~\cite{chard2020funcx}, a federated serverless framework. 
% Two real-world applications (i.e., weeds detection in precision agriculture and wildlife detection in camera traps) have been investigated.
\vic{Two real-world applications have been investigated: weeds detection in precision agriculture in Section~\ref{sec:weeds_detection} and wildlife detection in camera traps in Section~\ref{sec:camera_traps}.} 
In particular, unreliable clients with infrastructure-level errors will be studied in precision agriculture and ML-specific inconsistencies will be investigated in wildlife detection. \tl{To provide representative results, each scenario considers a very low number of clients (between 3 to 6).} Extensive experiments have been conducted and showed that even a simple FL algorithm is surprisingly robust to unreliable clients. \emph{By analysing FL robustness against unreliable clients, we aim to understand its impact in real-world applications. This will enable us to evaluate the level of sophistication required of FL algorithms for rural deployment. }

% Meanwhile, research works have been proposed to improve the fault tolerance of FL, but the impact of unreliable clients in real-world applications is not fully understood, especially in a rural environment which can introduces new challenges. For example, clients can become unavailable at any time due to unreliable power supply. The communication between clients and the server can experience long delay or even be impossible. Moreover, since the server does not govern the training process, clients may be misconfigured and generate a wrong result which may ``poison'' the shared model.  

\section{Problem Formulation}
%1. formulate the problem
%2. describe the algorithm we used in the experiments: For the weed detection one, we use Federated average. 
% specify we study two unreliable clients cases in the first application and the rest two in the second application
% emphasize the practicality of our problem, e.g., which application sees those problems

The two target scenarios belong to a classification problem with a set of $K$ data sources (or clients), all scattered over different locations. The statistical properties of each data source are not homogeneous, e.g. different class distributions. Each client is in charge of collecting and storing large amounts of local data. The traditional approach \cite{wang2021comparative}, named centralized approach in terms of data collection and model training, involves transferring all the data from each of the clients to a single server that will perform the training. This comes with a high network cost and potentially violates data sovereignty~\cite{nasr2019comprehensive}. 

To circumvent these drawbacks, FL proposes splitting the training between clients and the server \cite{yang2019federated}. The algorithm is fairly simple: it involves an infinite communication loop between the server and the clients. The server trains a generated (shared) model. A set of clients download a copy (weights) of the shared model and perform local updates leveraging private data, based on some optimization algorithm like stochastic gradient descent. Periodically, clients send a summary of the changes made to the local model (weights of the trained neural network) back to the server. Once it has gathered all the data, the server utilizes some aggregation techniques to perform an update to its shared model \cite{sannara2021federated}. 

\subsection*{Federated Averaging Algorithm} {A very popular aggregation technique is called \textbf{Federated Averaging} algorithm \cite{mcmahan17}. In this case, the server computes the average of the updates received from each client. The full pseudocode is depicted in Algorithm~\ref{alg:fedavg}. At the server, $K$ clients are selected, which are indexed by a variable $k$. In parallel, all clients update the generic model weights according to the \emph{ClientUpdate} function, which returns the trained weights $w$ back to the server. Finally, the server computes the average of all weights $w$ received from the $K$ clients. The average of the weights is regarded as the new set of weights for the generic model. Despite its clear advantages, Federated Averaging was not designed with specific fault tolerance features that might lead to sub-optimal model performance. Unreliable clients can be classified into two main types: those with infrastructure-level errors (e.g. faulty network), and the second involves ML-specific inconsistencies (flipped labels, misconfigured hyperparameters such as learning rate, other errors in collected data).
%those with malicious behavior and those dysfunctional behavior. }

\RestyleAlgo{ruled}
\DontPrintSemicolon 

\begin{algorithm}
\caption{\emph{FederatedAveraging}: The $K$ clients are indexed by $k$; $B$ is the local minibatch size, $E$ is the number of local epochs, and $\eta$ is the learning rate}\label{alg:fedavg}
%\KwData{$n \geq 0$}
%\KwResult{$y = x^n$}

%Source https://heartbeat.comet.ml/introduction-to-federated-learning-40eb122754a2
\SetKwFunction{FMain}{}
\SetKwProg{Sn}{Server executes}{:}{}
\Sn{\FMain{}}{
    initialize $w_{0}$\;
    \For{each round $t = 1,2, \ldots$}{
        $m \leftarrow max(C \cdot K,1)$\;
        $S_{t} \leftarrow$ (random set of $m$ clients)\;
        \For{each client $k \in S_{t}$ \textbf{in parallel}}{
            $w_{t+1}^k \leftarrow$ \textbf{ClientUpdate}($k$,$w_{t}$)\;
        }
        $w_{t+1} \leftarrow \sum_{k=1}^K \frac{n_k}{n}w_{t+1}^k$\;
    }   
}

%\BlankLine
\;
\SetKwProg{Cn}{Client update}{:}{}
\Cn{\FMain{$k$,$w$}}{
%//Run on client $k$
    $B \leftarrow$ (split $P_k$ into batches of size $B$)\;
    
    \For{each local epoch $i$ from 1 to $E$}{
        \For{batch $b \in B$}{
            $w \leftarrow w - \eta \nabla l(w;b)$\;
        }
    }
    
    return $w$ to server;
}

\end{algorithm}

% \subsection*{Robustness in Federated Learning} 
\subsection*{Infrastructure-level errors in Federated Learning} {These errors are due to uncontrollable events in the underlying system. During a global update cycle, the server assumes a total of $k$ weight updates, one belonging to each client. However, one or more clients might fail to send weight updates intermittently or permanently, due to a faulty network connection \cite{wang2022friends,deng2022secure}. Moreover, as stated earlier, each data source might show different statistical properties, such as class imbalances or even missing representation for certain class(es) \cite{wang2021addressing}. A faulty network coupled with heterogeneous data sources missing might lead to bad model performance, a model unable to generalize well to certain types of cases or even recognize certain classes at all. 

A real-world classification problem has been selected to assess the robustness of Federated Averaging learning towards such infrastructure-level issues. The target use case focuses on the detection of certain weed species from hyperspectral images. This kind of dataset requires large volumes of storage space, discouraging from transferring the data to a centralized location. It is the ideal candidate for Federated Learning-based training. The proposed set of experiments empirically analyzes the impact of network failures on the model accuracy, considering both permanent and intermittent failures from one or more clients simultaneously.}

\subsection*{ML-specific inconsistencies in Federated Learning} {Clients with ML-specific inconsistencies disrupt the model training process, compromise its accuracy or even change the model behavior altogether. The central model relies on properly labelled data from each data source. The primary source of irregularities might come from the data itself: Certain clients might provide mislabelled training data or intentionally flipped class labels \cite{lv2022awfc,jebreel2022defending}. As a result, the resulting model will not be able to detect certain class(es), deeming the system unusable. 

Federated learning relies on a number of configuration parameters (see Algorithm~\ref{alg:fedavg}) to tune the ultimate model accuracy. In certain situations, some clients may have wrong values of such parameters with malicious intentions or simply due to a misconfiguration and will degrade the model performance. The learning rate $\eta$ has been shown to be a parameter of choice \cite{lv2022awfc,jebreel2022defending} and able to affect the performance of both the aggregated, central model and the local ones ~\cite{bagdasaryan20}.

In order to evaluate the effects of ML-specific inconsistencies in federated learning, a second real-world application has been utilized, named animal detection from trap cameras. Such trap cameras are placed on different locations and collect bursts of (at least) three images whenever the motion sensor is triggered. The experiments are designed to analyze the separate effects of mislabelled data and incorrect learning rate values, and also explore whether the combination of both kinds (flipped labels and learning rate values) amplifies the negative effect.
}

As explained above, Federated Averaging Learning was not designed with fault tolerance behavior. Intuitively, we would expect the model performance to degrade as infrastructure-level or ML- specific inconsistencies develop. The experiment set carried out through two real-world applications will prove (empirically) that Federated Average Learning has a bigger than expected tolerance for such behavior. 

\section{FL in Precision weeds detection}\label{sec:weeds_detection}

\begin{figure*}[t]
    \centering
    \includegraphics[width=0.8\linewidth]{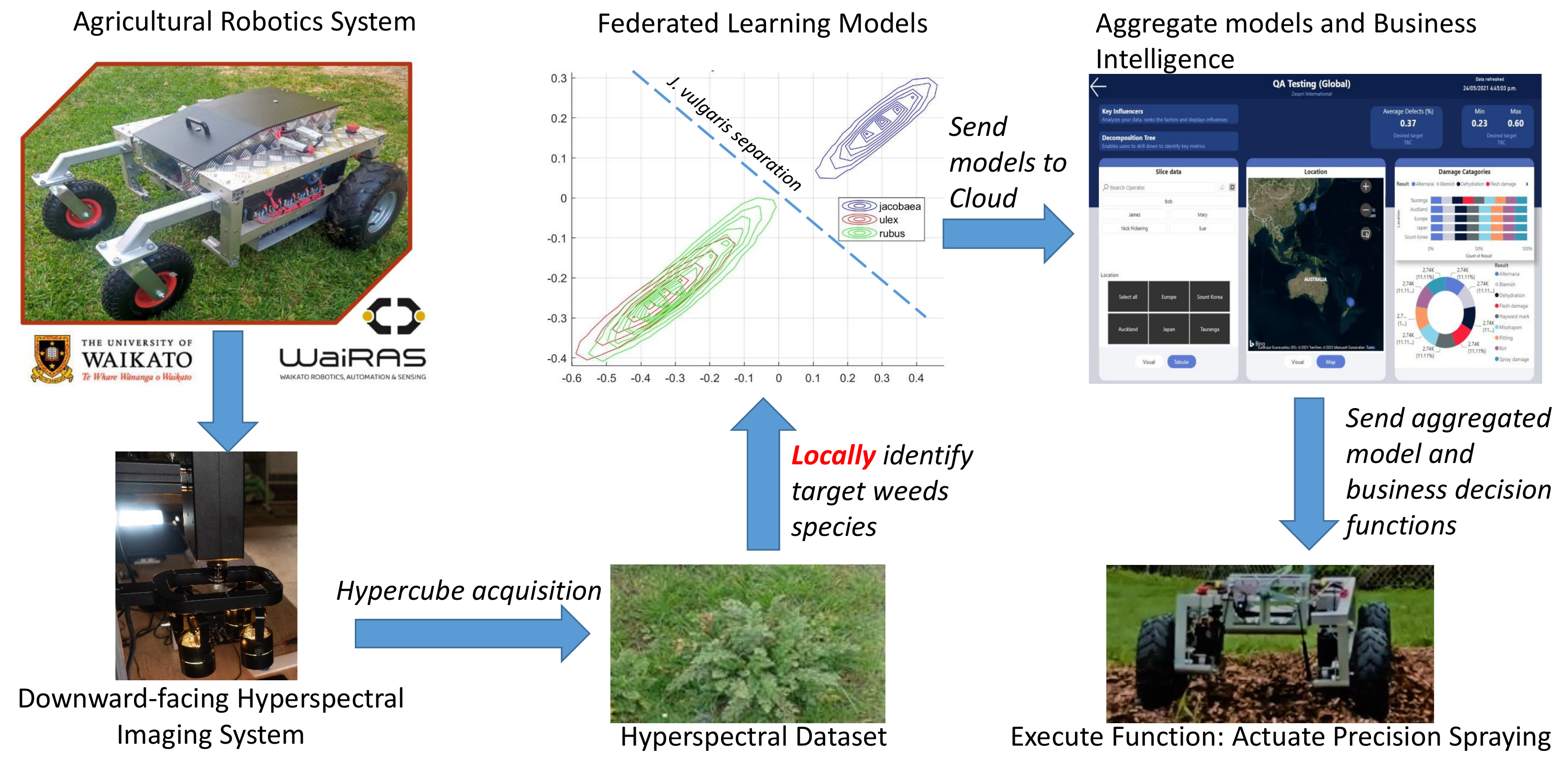}
    \caption{Hyperspectral Weeds Detection System}
    \label{fig:NIR-HSI}
\end{figure*}

We focus on precision weeds detection as our rural agriculture application. This robotic system uses a near-infrared hyperspectral imaging system to guide the operations of a mobile precision pesticide spraying system~\cite{holmes2019pasture} (see \figurename~\ref{fig:NIR-HSI}). The use of near-infrared hyperspectral cameras as a replacement for standard RGB cameras is motivated by research suggesting that spectral measurements can more accurately distinguish weeds from produce~\cite{holmes2019pasture}.
However, the more sophisticated cameras also increase power and data requirements, leading to a need to optimize power and energy consumption in order to achieve maximum area coverage.
An effective solution should process camera data locally for weed detection, but also communicate a locally processed model to the global database for aggregation. Therefore, FL is a good fit for this application. 

Our intent is to investigate the practical aspects of FL for weeds detection where we deploy a Pasture Care Robot with hyperspectral imaging capabilities. Note that in a large pasture area, full network coverage may not be guaranteed. When the robot is operating in the pasture, it is unavoidable that the robot will completely drop out during the FL training due to network connection issues. Moreover, when the robot's power level is low, it is a common engineering practice that the robot will prioritize navigation over data transmission. In other words, the robot may still receive data but not transmit data in order to conserve battery. Motivated by these, in this weeds detection application, we will investigate the unreliable client scenarios where clients completely or partially drop in and out of the federation. 
% 1. Pastue area is large, not all will be under a network coverage
% 2. Power functions have priority, once the power level is low, robot will shut transmission and prioritise energy for navigation 

%Therefore, the chosen dataset and designed experiments are intended to provide real-world deployment and engineering requirements, such as accuracy and robustness. 

\subsection{Dataset} \label{subsec:weed_dataset}
As summarized in Table~\ref{tab:dataset}, the dataset we used in our experiments consists of hyperspectral pasture images taken from three different rural sites~\cite{holmes2019pasture}. The dataset was labelled with four classes including three different species of pastoral weed: \emph{Jacobaea Vulgaris} (common name: \emph{Ragwort}), \emph{Ulex} (common name: \emph{Gorse}), \emph{Rubus} (common name: \emph{Blackberry}), and a background class of grass. These near-infrared hyperspectral images were taken using a Pika hyperspectral camera and normalised against a white reference to convert from radiance to reflectance while also removing instrument-specific variations. Sample points were extracted from the image datacubes, whereby each pixel had 148 dimensions in the infrared spectrum of 900nm to 1700nm. The dataset was partitioned into training (80\% of the samples) and testing (20\% of the samples) datasets in a randomized class-stratified manner. 

Class normalization of dataset was not performed as our evaluation considers each hyperspectral image individually. In a live environment, it is conceivable that a site has insufficient examples of any given class such that normalization would remove large portions of valid data. We consider this data valid as we compare federated and centralized models to local models: models that would use such data exclusively. A fundamental part of the evaluation of FL in this case is the ability to benefit from the knowledge of other sites.

\begin{table}[]
\centering
    \caption{Hyperspectral pasture image dataset with imbalanced class distributions and disparate volumes of data among different sites (W: Pastoral weeds, G: Background grass)}\label{tab:dataset}
    \begin{tabular}{l|ccc|c}
    Location & Site A & Site B & Site C & Total \\ \hline
    \#Samples & 60,072 & 30,240 & 6,232 & 104,544 \\
    \#Classes & 4 (3W + G) & 4 (3W + G) & 2 (1W + G) & 4 (3W + G) \\
    \hline 
    
    \hline
\end{tabular}
\end{table}

Note that the heterogeneity in our dataset poses challenges for FL: (1) \emph{Data heterogeneity on class labels}. As shown in Table~\ref{tab:dataset}, data from the first two sites were balanced across all four of the classes while data from the third consisted of only one weed species and grass, making the distribution of examples between sites quite different and therefore potentially challenging for FL. 
(2) \emph{Data heterogeneity on data volumes}. 
% An additional challenge was the disparate volumes of data from the three sides: 
Site A provided approximately two times as much data as site B and ten times as much data as site C.

\subsection{Experiment Plan}
% Todo (Vic): summarize the objectives in a table (check Proximal Near-Infrared Spectral Reflectance Characterisation of Weeds Species in New Zealand Pasture)

\begin{table*}[]
\centering
\caption{Experiment plan}
\label{tab:exp_plan}
\begin{tabularx}{\linewidth}{p{1.5cm}|p{6cm}|p{2cm}|p{4cm}|X}%{X|X|X|X|X}
\textbf{Objective} & \textbf{Research question} & \textbf{Algorithm} & \textbf{Independent variables} & \textbf{Dependent variables} \\ \hline
Classification accuracy & How well does FL perform compared with other ML approaches? & FL, Centralized, Localized & Hyperparameters (e.g., iterations, minibatch size, \#epochs/iteration) &  \\ \cline{1-4}
& What is the impact of clients partially dropping out on FL? &  & When the clients drop out (e.g., upload, download) &  \multirow{4}{*}{Classification accuracy}\\ \cline{2-2} \cline{4-4}
\multirow{3}{*}{Robustness} & What is the effect of training data volume on FL with unreliable clients? & \multirow{3}{*}{FL} & Which client/site drops out (sites with different data volume) &  \\ \cline{2-2} \cline{4-4}
 & What is the effect of client participation rate on FL? &  & Client participation rate &  \\ \cline{1-2} \cline{4-5}
% Resource consumption & How much resources does an edge device need to run FL? &  & NA & 
% \begin{tabular}[c]{@{}l@{}}Memory, CPU utilization,\\ network bandwidth\end{tabular} \\ \hline   
\hline
\end{tabularx}
\end{table*}

We evaluate the capabilities of FL in supporting cross-field hyperspectral weeds detection from two aspects:

\begin{figure}
    \centering
    \includegraphics[width=0.7\linewidth]{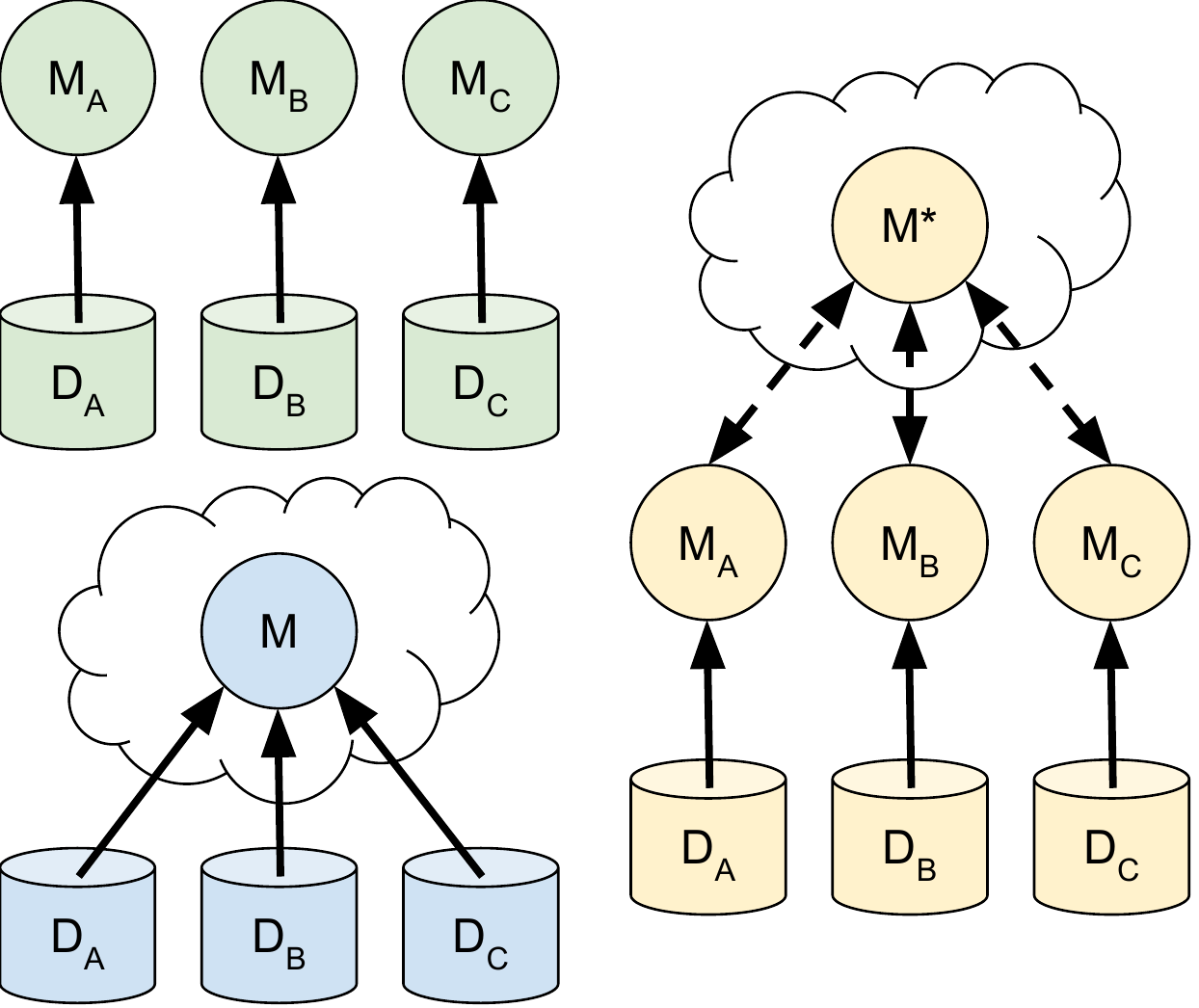}
    \caption{Approaches to ML in rural AI. Green illustrates a typical fully localised system where data is not shared between sites;
    each model is locally relevant only 
    with no communication between sites.
    Blue illustrates a fully centralised system where all data is uploaded to the cloud and a single centralised model is trained;
    while potentially the most accurate approach,
    this cannot be used with sensitive data or where the amount of data to be transferred is too large for the available bandwidth.
    Yellow shows a federated approach where data is kept local while models are sent to the parameter server
    and used to compute an aggregate model which in turn
    is shared back to the local sites;
    this allows learning across sites
    while not requiring data to be shared.}
    \label{fig:ML-approaches}
\end{figure}

\subsubsection{Classification accuracy} 
To justify the use of FL, we compared FL with a fully localized approach and a fully centralized approach as shown in Fig.~\ref{fig:ML-approaches}. The fully localized approach (Fig.~\ref{fig:ML-approaches}, green figure) represents a completely decoupled system where each model is trained locally at the three sites and there is no communication at all between sites.
The fully centralized approach (Fig.~\ref{fig:ML-approaches}, blue figure) represents a completely centralized approach where the model resides on a cloud server and all data must be uploaded to the server before the model is trained.
These two approaches represent the current alternative status quo, both of which have disadvantages. For example, the centralized approach can introduce large communication overhead because it requires all data to be uploaded to the cloud. Moreover, the centralized approach cannot be used with sensitive data due to the violation of data sovereignty. Although these issues can be addressed by the localized approach, a model trained using local data only may not perform well, especially with a heterogeneous dataset. 

On the other hand, in FL (Fig.~\ref{fig:ML-approaches}, yellow figure), data resides locally and only the local models are transferred to the cloud server. The cloud server aggregates the models and then returns them back to the local sites. In this way, information and knowledge are shared between the sites indirectly by way of model sharing, as opposed to directly by way of data uploading.
It can be expected that the centralised approach will be the most accurate due to complete access to the data and that the FL approach will be superior to the local-only approach.

\subsubsection{Robustness against unreliable clients}
In rural agricultural farmlands, edge devices (i.e., clients) may sometimes experience unreliable power supply or network connection, which leads to devices dropping in and out during the training process. As a result, a trained model from a local site may not be uploaded to the cloud server for model aggregation. Alternatively, when the failure happens during the model downloading process, the local model is not updated with the aggregated model. To demonstrate the performance of FL for withstanding node failure, we simulate different scenarios under different client participation rates ranging from 100\% to 25\%. First of all, we investigate the impact of training data volume under the unreliable client setting on the performance of FL. Second, we evaluate how the failures during the model uploading or downloading process affect the performance of FL.
% sample from a uniform distribution for a device in each training epoch. A device will participate the training only if the sampled number is larger than the pre-defined failure rate.  

\subsection{Results and Discussion}

\subsubsection{Comparison of different ML approaches}
\hfill\\
The first six columns of Table \ref{tab:bestaccuracy} 
give a summary of our hyperparameter optimisation results for the short 200 iteration runs. It can be immediately observed that the local models have significantly lower accuracies, with the model at site A achieving a best-case accuracy of marginally above 70\%, and the model at site C achieving, at best, 39.5\% accuracy. On the other hand, the centralized model achieves nearly 99\% test accuracy.

\begin{figure}
    \centering
    \includegraphics[width=0.8\linewidth]{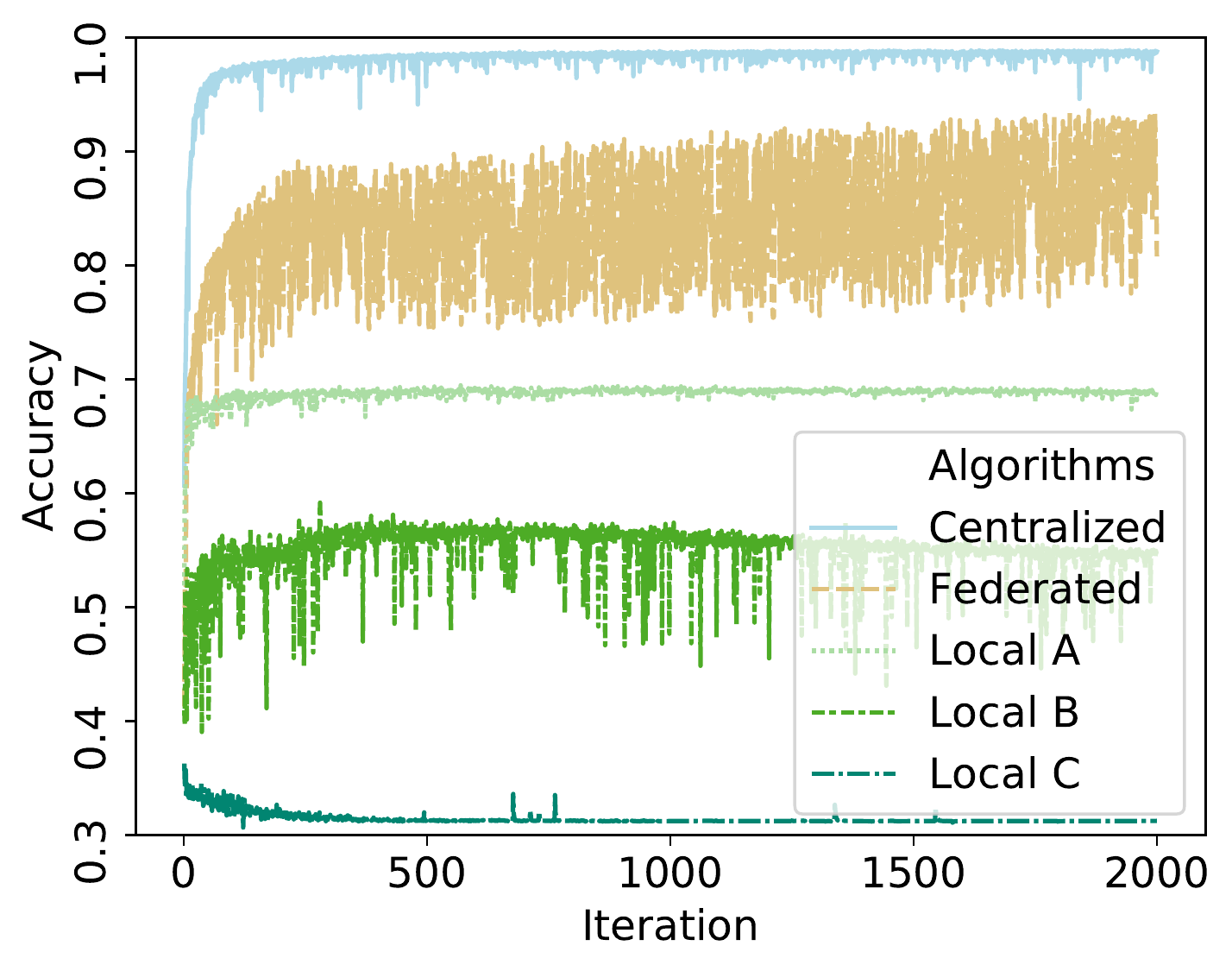}
    \caption{A comparison between FL, the centralized approach, and the local approach (see Fig.~\ref{fig:ML-approaches}) to prove why we should use FL when the centralized approach is not applicable (e.g., high communication cost).}
    \label{fig:ml_comparison_result}
\end{figure}

\begin{table*}[]
    \centering
    \caption{Best testing accuracies by hyperparameter combination (columns) and model approach as shown in Fig.~\ref{fig:ML-approaches} (rows). Where $T=200$, the best accuracies for each approach are bolded. Where $T\geq2,000$, the accuracy for an approach is bolded only if it exceeds the best accuracy in the preceding columns.
    }
    \label{tab:bestaccuracy}
    \begin{tabular}{l|cccccc|cc}
         Model  & \multicolumn{7}{c}{(\#Epochs/Iteration $E$, Minibatch Size $B$, Iterations $T$)}\\
         Type   &(1,10,200)&(5,10,200)&(20,10,200)&(1,50,200)&(5,50,200)&(20,50,200)&(1,50,2000)&(1,50,10000)\\
         \hline
         Local (A) &0.693	&\textbf{0.701}	&0.695	&0.692	&\textbf{0.701}	&0.699&0.695&0.697\\
         Local (B) &0.571	&0.576	&0.555	&0.569	&0.590&0.577&0.592&\textbf{0.601}\\
         Local (C) &\textbf{0.395}	&0.341	&0.321	&0.382	&0.346	&0.339&0.363&0.378\\
         Centralized &0.985	&\textbf{0.988}	&\textbf{0.988}	&0.980	&0.987	&\textbf{0.988}&\textbf{0.989}&\textbf{0.989}\\
         Federated   &0.840	&0.809	&0.837	&\textbf{0.883}	&0.863	&0.861&0.936&\textbf{0.963}\\
        \hline
    
        \hline
    \end{tabular}
\end{table*}

The table also shows the accuracy of the best federated model. This approach, after 200 iterations, reaches the best performance of just over 88\% when the number of epochs per
iteration is low ($E=1$) and the batch size is higher ($B=50$).
Unlike the centralized model, however, the federated approach is relatively sensitive to the choice of hyperparameters: specifically, $E=1$ and $B=50$ appear to be good choices and we proceeded with further experiments using these hyperparameters.

\begin{figure*}[!tb]
      \begin{center}
        \subfloat[Unreliable Client A]{\includegraphics[width=0.33\linewidth]{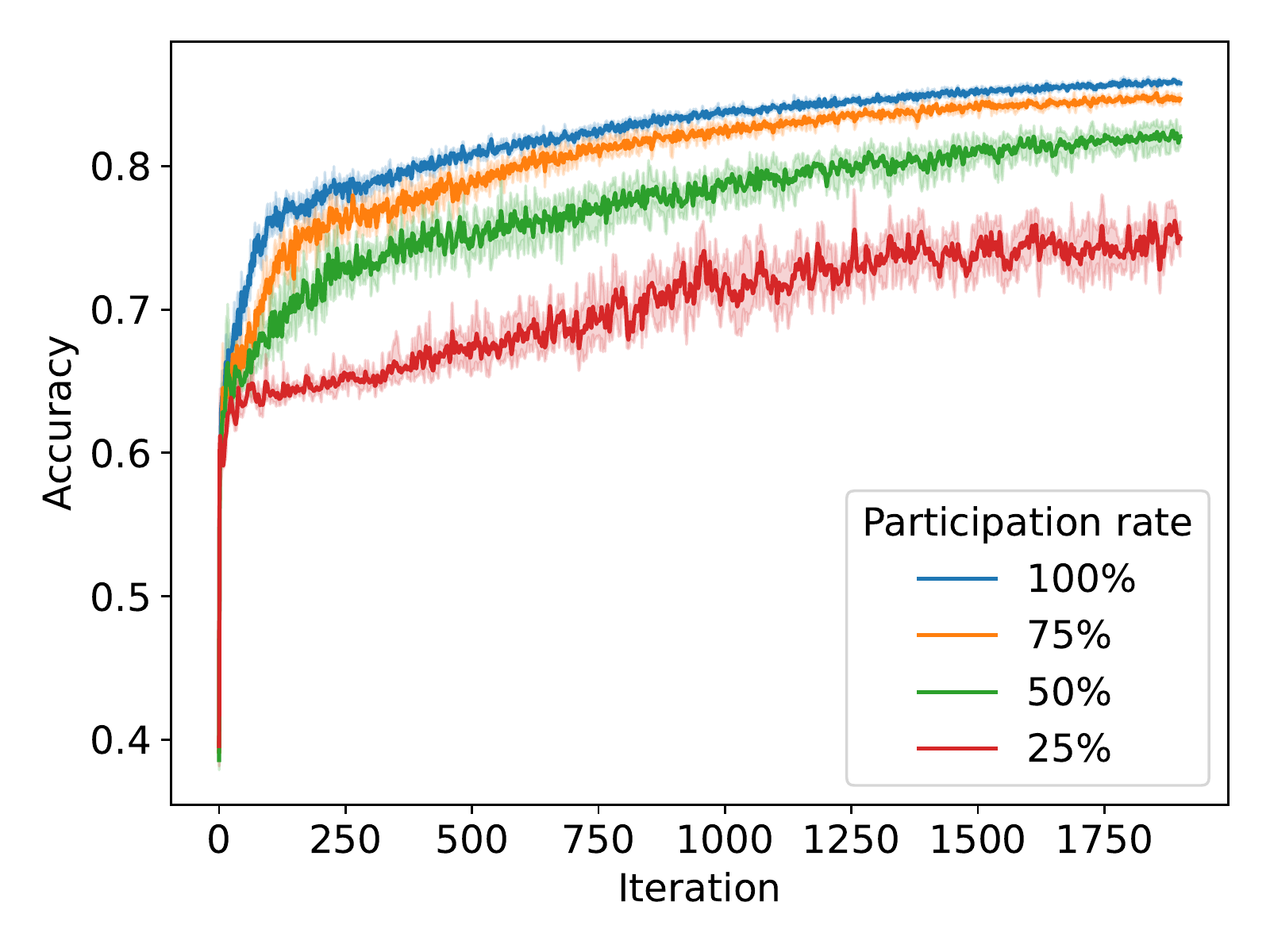}} %\enspace
        \subfloat[Unreliable Client B]{\includegraphics[width=0.33\linewidth]{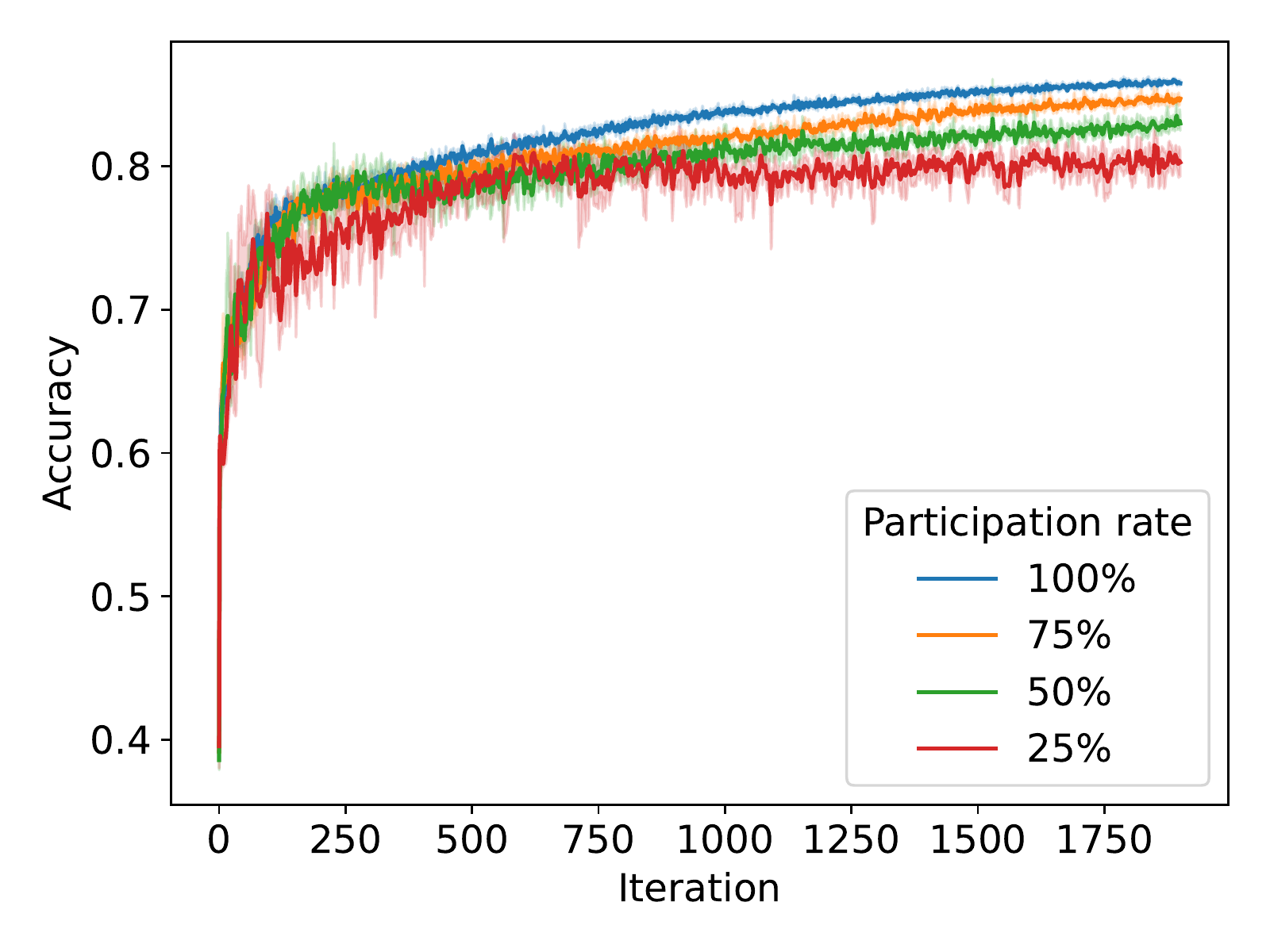}}
        \subfloat[Unreliable Client C]{\includegraphics[width=0.33\linewidth]{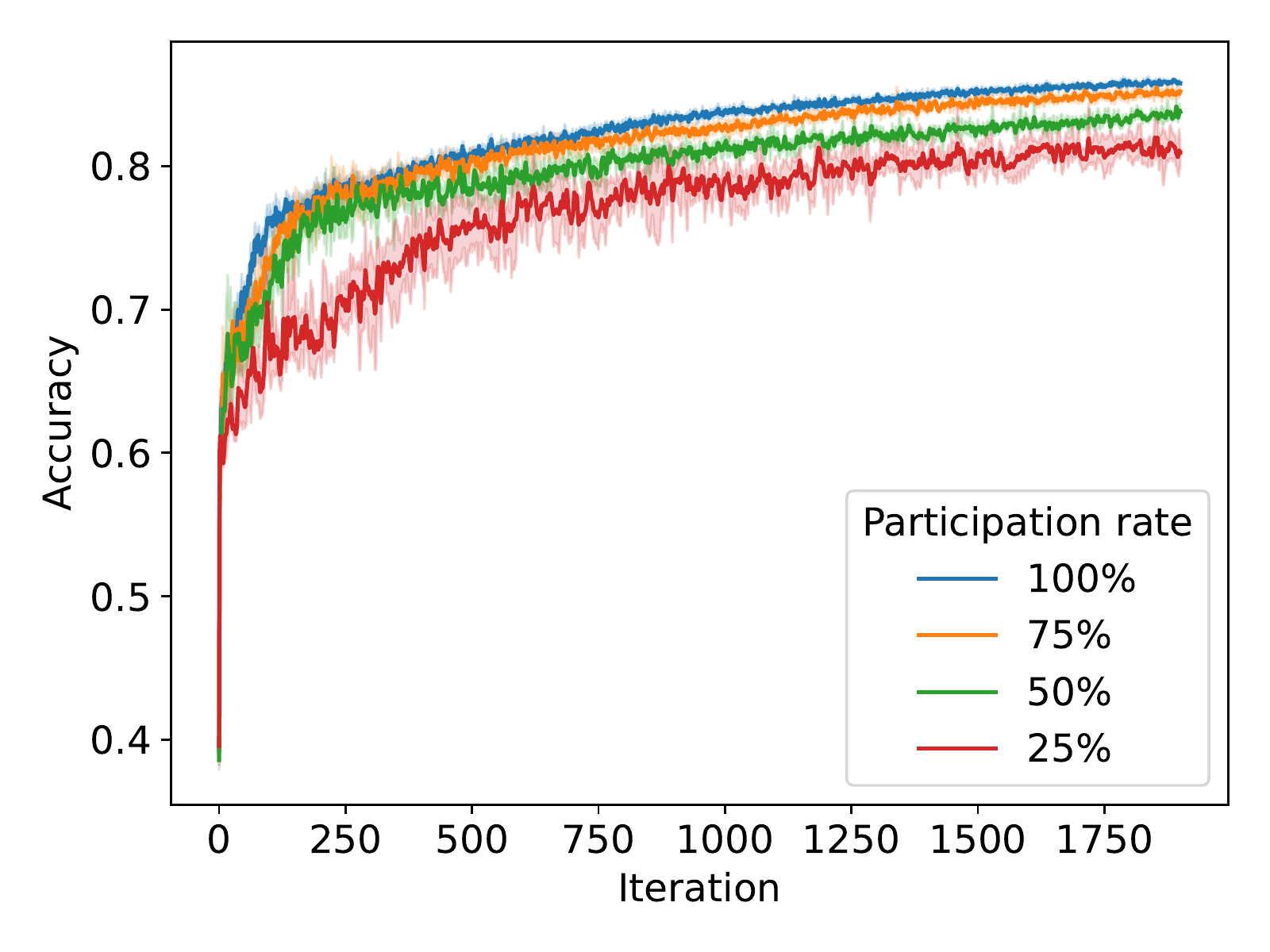}}
      \end{center} 
    \caption{The training performance of FL with random and complete dropout clients. Each sub-figure shows the FL training performance when a site randomly and completely drops out during FL training process with various probabilities.}
    \label{fig:fl_unreliable_download_upload}
\end{figure*} 

\begin{figure*}[!tb]
      \begin{center}
        \subfloat[Unreliable upload]{\includegraphics[width=0.35\linewidth]{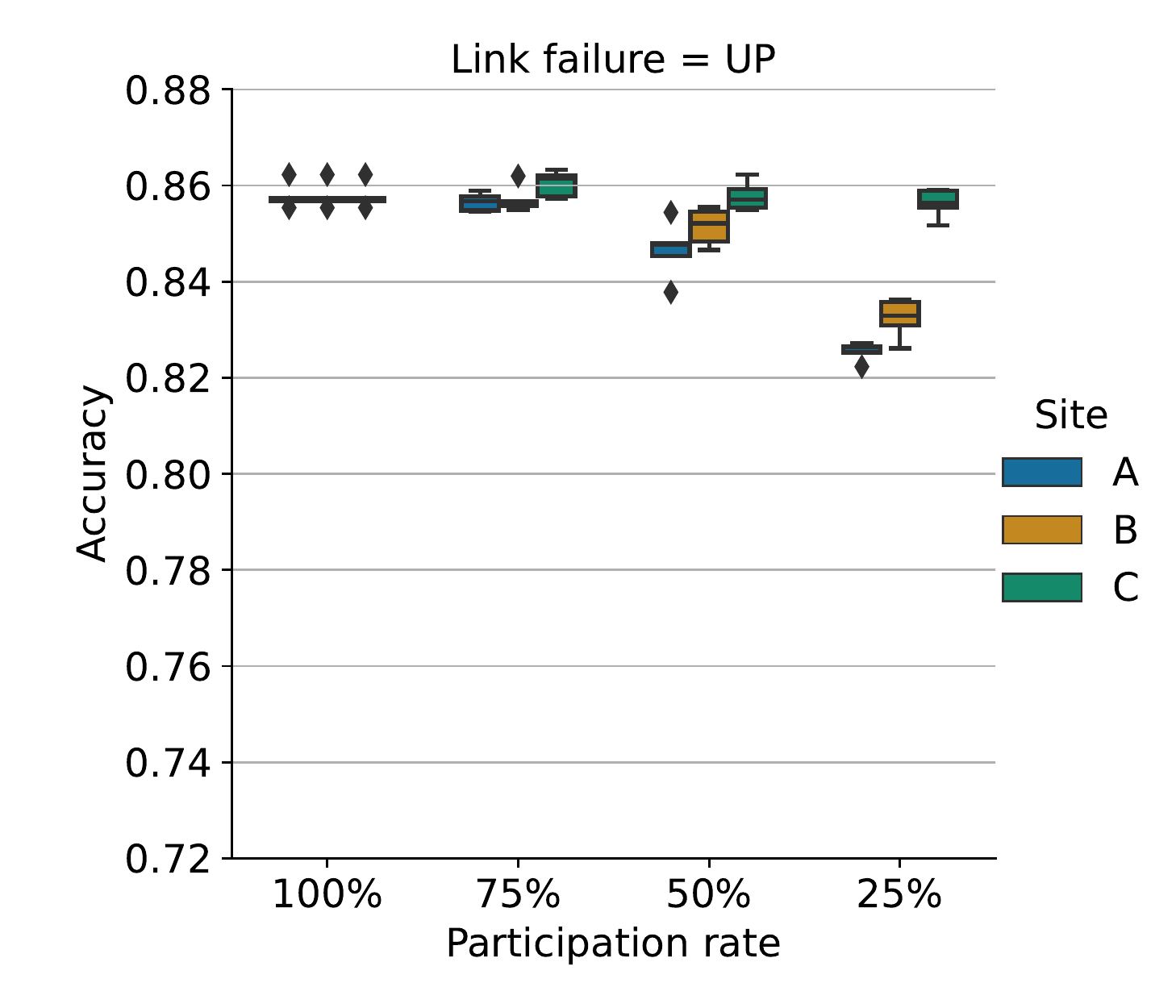}} %\enspace
        \subfloat[Unreliable download]{\includegraphics[width=0.35\linewidth]{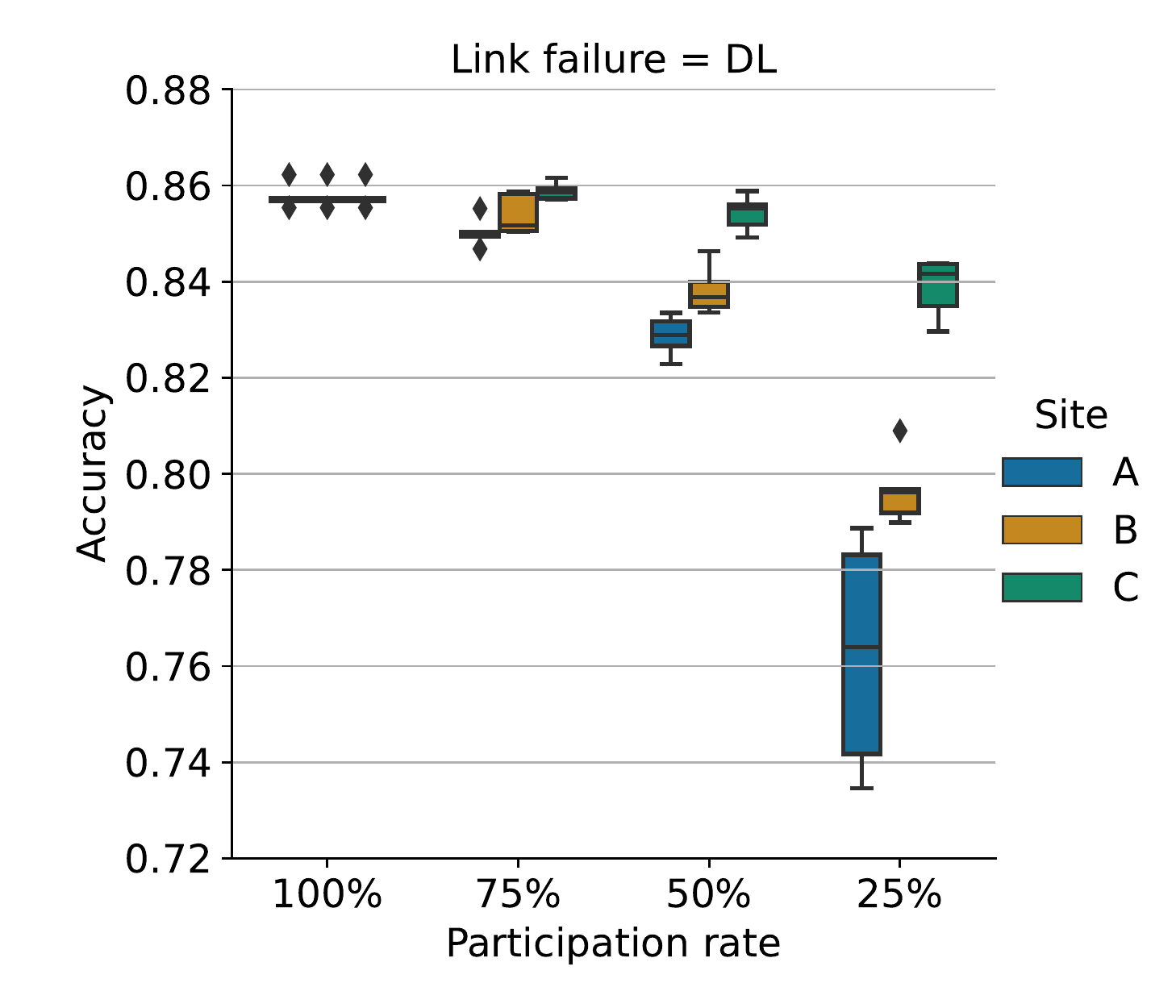}}
      \end{center} 
    \caption{FL model accuracy with random and partial dropout clients. Note that we select the model after 2,000 training iterations. Each sub-figure demonstrates that under the same link failure, how data heterogeneity affects the FL accuracy. ``UP'' and ``DL'' indicate unreliable model upload and download respectively. From both (a) and (b), we can see that site A has the highest impact of the overall accuracy due to its largest data volume and class labels.}
    \label{fig:fl_diff_failure_at_site}
\end{figure*} 

\begin{figure*}[!tb]
      \begin{center}
        \subfloat[Unreliable Client A]{\includegraphics[width=0.33\linewidth]{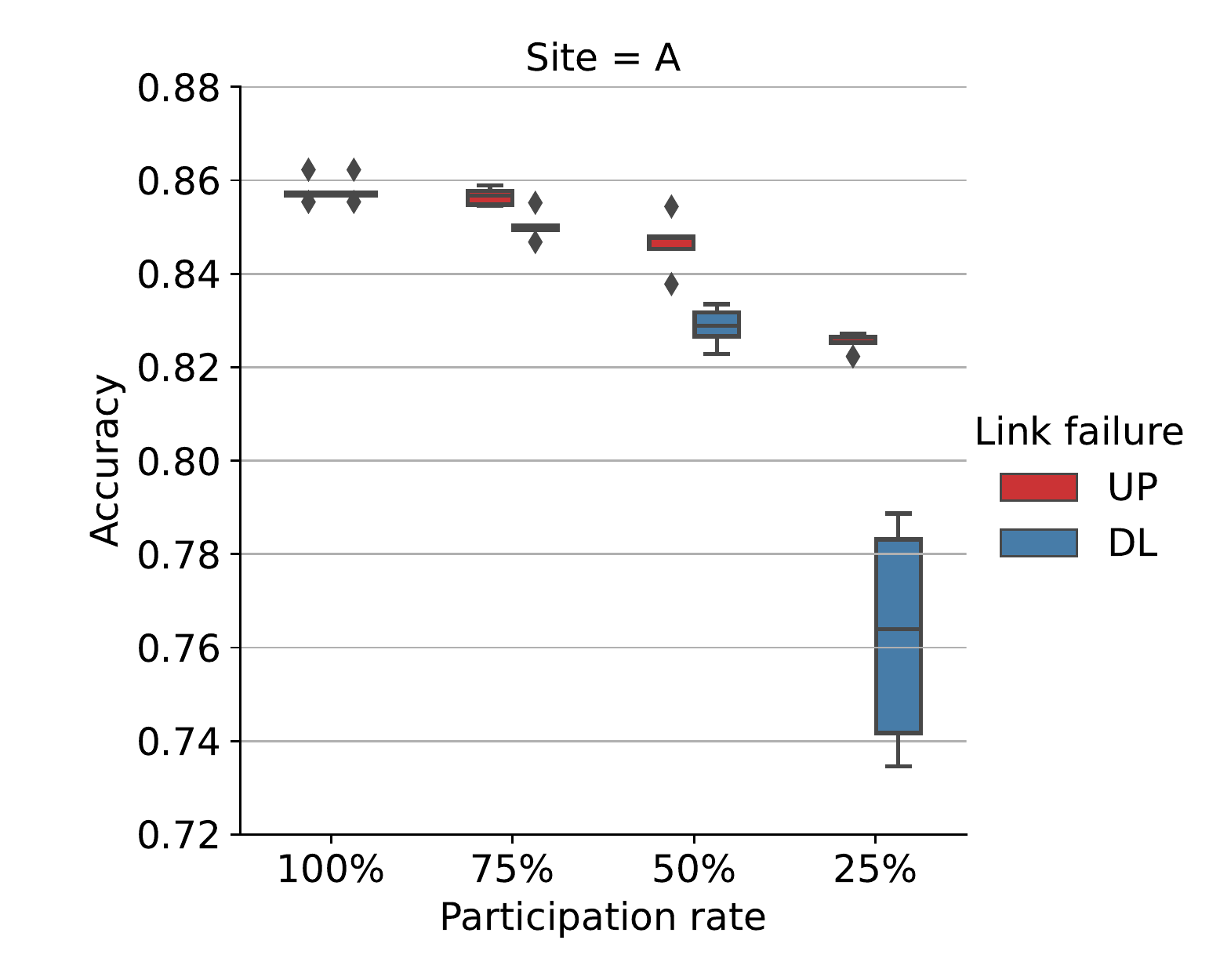}} %\enspace
        \subfloat[Unreliable Client B]{\includegraphics[width=0.33\linewidth]{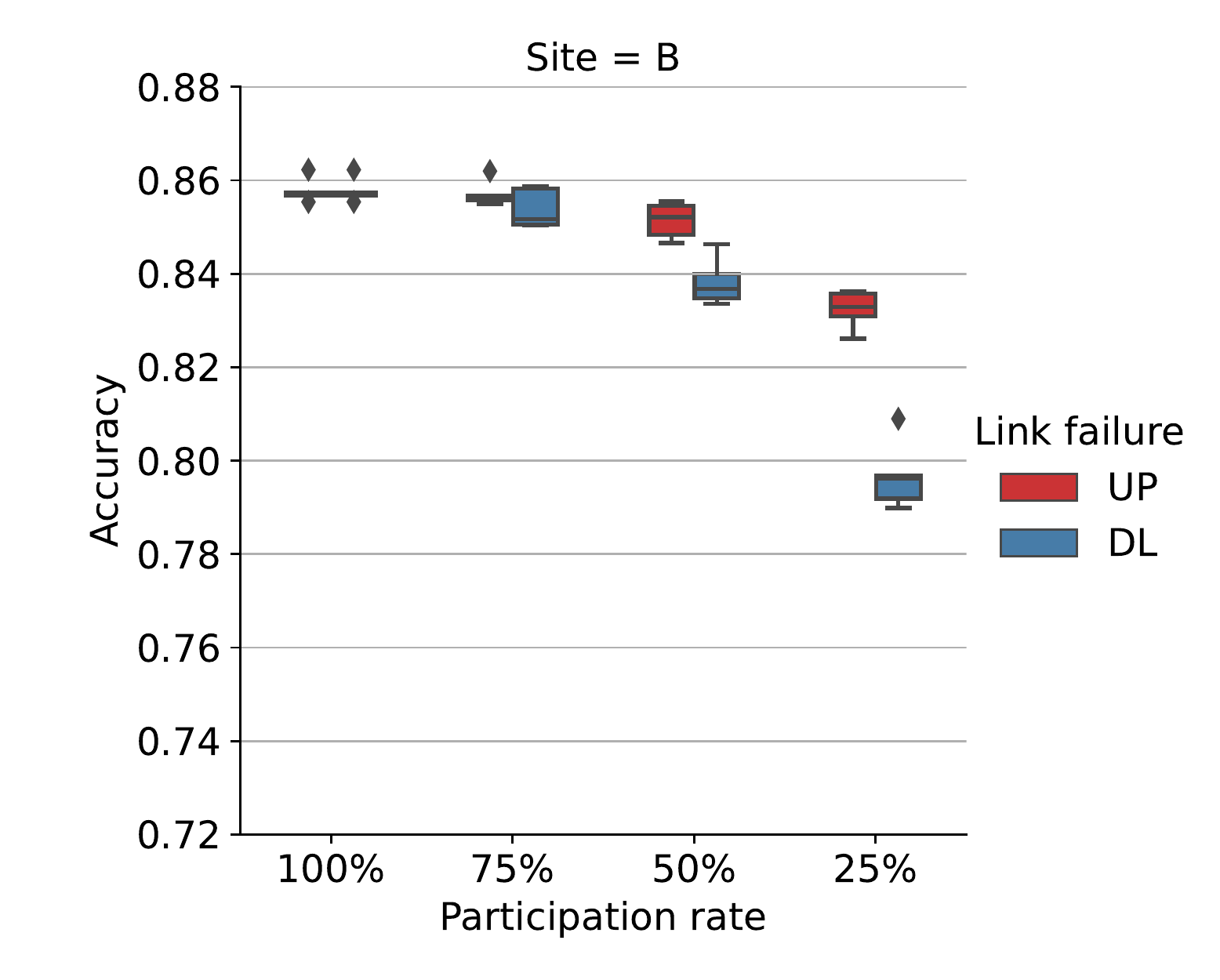}}
        \subfloat[Unreliable Client C]{\includegraphics[width=0.33\linewidth]{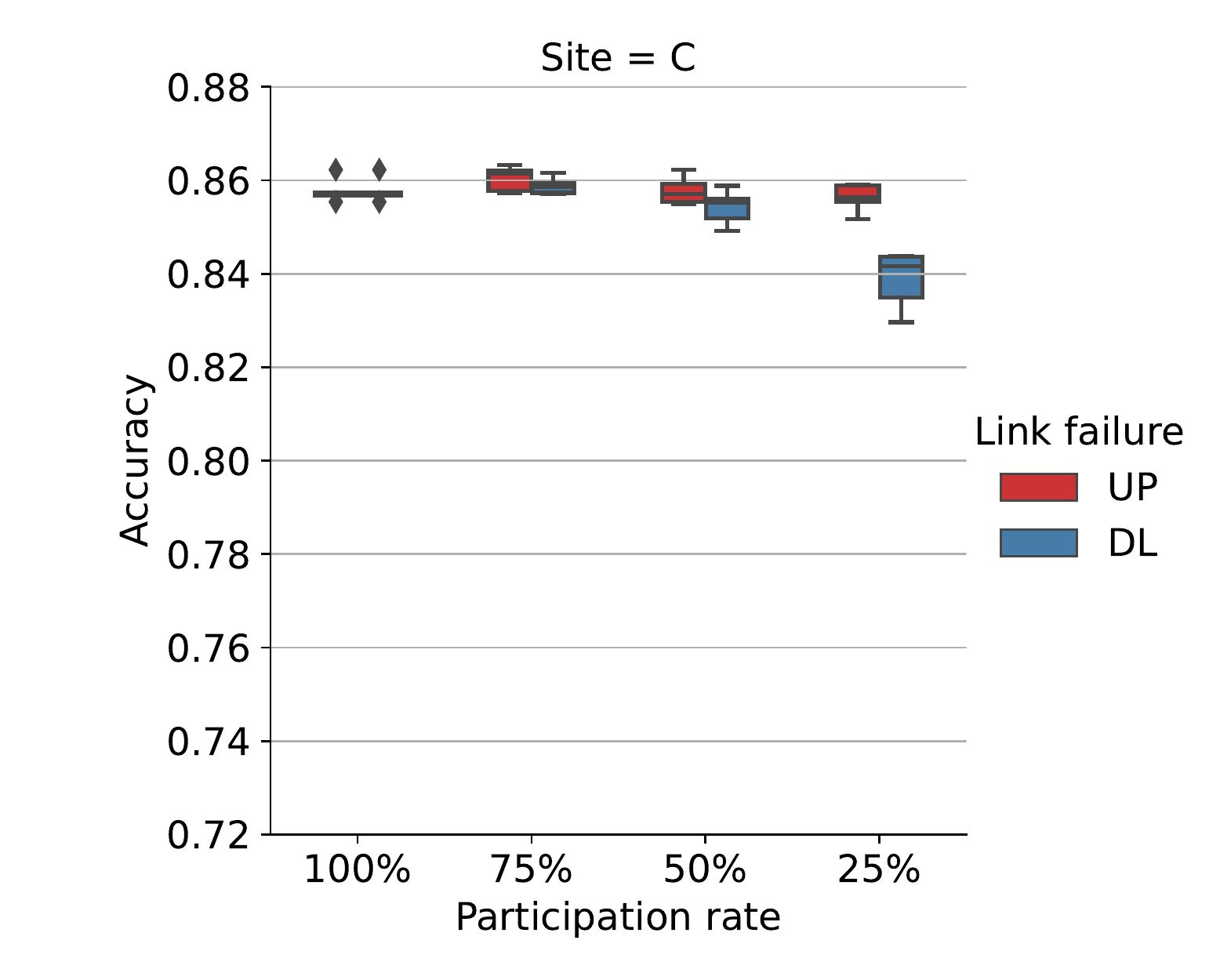}}
      \end{center} 
    \caption{Performance comparison between different clients and dropout types. Each sub-figure demonstrates that within the same site (i.e., the same training dataset), how different failure types affect the FL accuracy. ``UP'' and ``DL'' indicate unreliable model upload and download respectively. For example, from all 3 figures, upload failure barely affects the FL accuracy compared to download failure.}
    \label{fig:fl_diff_site_failure}
\end{figure*}

The results of the further experiments are also given in Table~\ref{tab:bestaccuracy} (rightmost two columns) and also Fig.~\ref{fig:ml_comparison_result}. This column and the figure show that the federated model significantly improves its accuracy with additional iterations. 
The best model found with FL achieves nearly 94\% test accuracy after 2,000 iterations and exceeds 96\% accuracy after 10,000 iterations, both good improvements compared to the 200-iteration case. Despite its lack of centralised data, FL can still achieve 96\% accuracy as the centralized approach, demonstrating that FL will be a good fit for applications where centralized training is inapplicable. Note that in both cases, the models labelled as ``local'' are not part of the federation and simulate a situation where models train locally and there is no information sharing between sites at all.
% The best model found with FL achieves nearly 94\% test accuracy after 2,000 iterations and exceeds 96\% accuracy after 10,000 iterations, both good improvements compared to the 200 iteration case, and illustrating the fact that lack of centralised data in FL means that convergence of the model is slower. Note that in both cases, the models labelled as ``local'' are not part of the federation and simulate a situation where models train locally and there is no information sharing between sites at all.

\subsubsection{FL robustness against client dropout}
\hfill\\
\textbf{Clients completely drop out.}
Fig.~\ref{fig:fl_unreliable_download_upload} shows the training performance of FL when different clients randomly and completely drop out. For example, in Fig.~\ref{fig:fl_unreliable_download_upload}(a), Client A will randomly drop out during the training based on a pre-defined participation rate. From Fig.~\ref{fig:fl_unreliable_download_upload}(a), we can see that the training performance is barely affected when Client A only participates 75\% of the training iteration. However, larger gaps can be observed when the participation rate drops to 50\% or lower. Nevertheless, even with a 75\% decrease in participation rate compared to 100\% participation, only less than 15\% reduction in model accuracy is observed. This result confirms that FL is robust to random client dropout. 

The impact of different clients dropout can also be seen by comparing Fig.~\ref{fig:fl_unreliable_download_upload}(a), Fig.~\ref{fig:fl_unreliable_download_upload}(b), and Fig.~\ref{fig:fl_unreliable_download_upload}(c). Due to the heterogeneity in our dataset (as we mentioned in Sec.~\ref{subsec:weed_dataset}), the impact of different sites dropping out on FL model accuracy varies. For example, Client A has a more significant impact where almost a 10\% accuracy decrease can be spotted with a participation rate decrease from 50\% to 25\%. On the other hand, we can only see less than a 5\% decrease in accuracy when Client B or C drops out. As we expected, the influence of a client on FL depends on the amount of data and the number of classes it has.

\textbf{Clients partially drop out.} We define partial dropout as clients may randomly fail to upload their local model or download the aggregated model during model training. 
Fig.~\ref{fig:fl_diff_failure_at_site} and Fig.~\ref{fig:fl_diff_site_failure} compared the performance of the models trained at various client participation rates. 

Fig.~\ref{fig:fl_diff_failure_at_site}(a) shows the impact of clients randomly failing to upload their locally trained models. We can see that there is less than a 4\% accuracy decrease when a client does not upload its model. Meanwhile, we also notice that the negative impact is less noticeable if the dropout happens in the client with fewer data, which matches our previous observation. Similar conclusions can also be drawn in Fig.~\ref{fig:fl_diff_failure_at_site}(b) when a client fails to download the aggregated models.

In Fig.~\ref{fig:fl_diff_site_failure}, we compared the different impacts of failing in upload or download on the FL model accuracy. For example, we can see from Fig.~\ref{fig:fl_diff_site_failure}(a) that clients dropping out during model download has a larger impact on the model accuracy. This observation holds for different participation rates as well as different clients.

\begin{figure*}
     \centering
     \begin{subfigure}[b]{0.23\textwidth}
         \centering
         \includegraphics[width=1.0\linewidth]{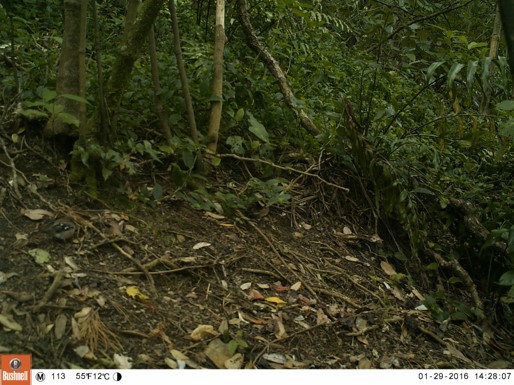}
         \caption{True Positive}
         \label{fig:a}
     \end{subfigure}
     \hfill
     \begin{subfigure}[b]{0.23\textwidth}
         \centering
         \includegraphics[width=1.0\linewidth]{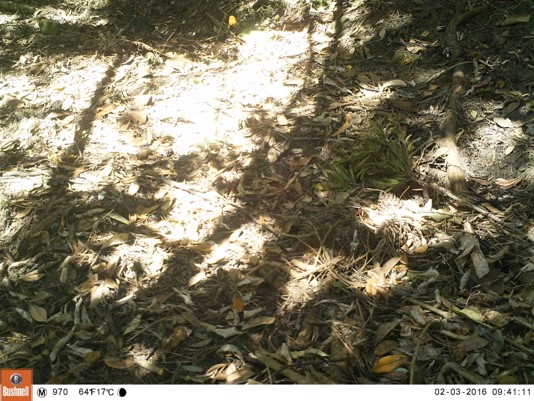}
         \caption{Possible False Positive}
         \label{fig:b}
     \end{subfigure}
     \hfill
     \begin{subfigure}[b]{0.23\textwidth}
         \centering
         \includegraphics[width=1.0\linewidth]{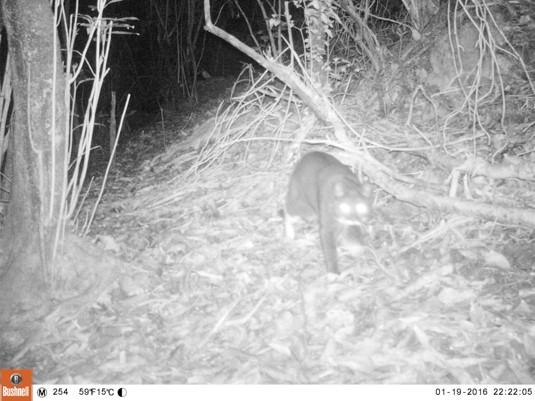}
         \caption{Night Mode}
         \label{fig:c}
     \end{subfigure}
     \hfill
     \begin{subfigure}[b]{0.23\textwidth}
         \centering
         \includegraphics[width=1.0\linewidth]{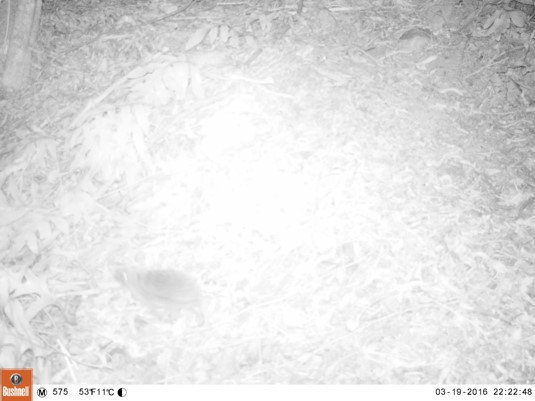}
         \caption{Over Exposed Night Mode}
         \label{fig:d}
     \end{subfigure}
        \caption{Examples of single images from Wellington camera trap dataset}
        \label{fig:ExampleImages}
\end{figure*}

\begin{figure*}[!tb]
      \begin{center}
       {\includegraphics[width=0.95\linewidth]{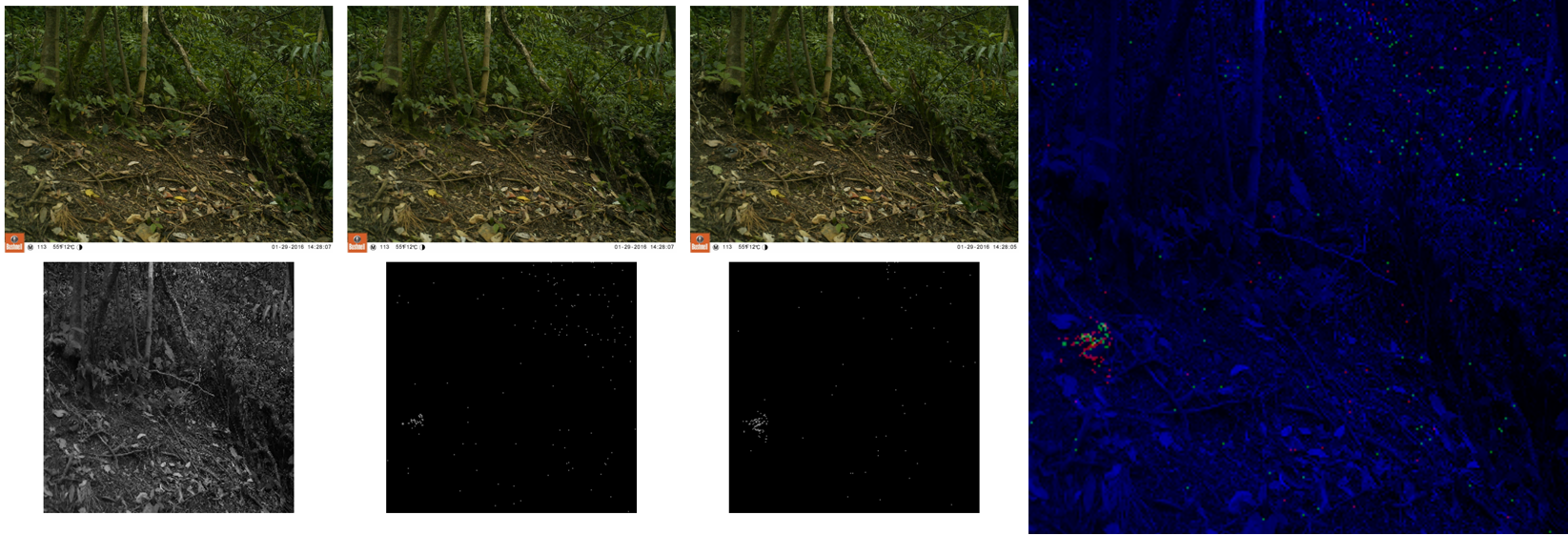}} %\enspace
      \end{center} 
    \caption{Preprocessing of image bursts}
    \label{fig:Preprocessing}
\end{figure*} 

\section{FL in Camera traps} \label{sec:camera_traps}
The current biodiversity crisis requires a global network of remote cameras to monitor and measure essential biodiversity variables using non-invasive remote cameras \cite{steenweg2017scaling}. A camera trap is a camera with a motion sensor and infrared flash suitable for sampling medium and large-sized birds and mammals with minimal disruption to animal behaviours. There is an increasing trend of using camera traps for monitoring different biodiversity trends, for example, monitoring highway crossing for observing the impact of human activities on wildlife~\cite{barrueto2014anthropogenic}, and studying the impact of climate change, population shape, and trophic interactions on elk \cite{brodie2014trophic}. Recently, ML has been very successfully used in image-based biodiversity surveys for decision-making purposes by answering critical conservation questions \cite{zualkernan2022iot,pantazis2021focus,norouzzadeh2021deep}. However, there are limited coordinated efforts among these camera studies \cite{steenweg2017scaling}. One limitation is the hesitation in sharing data among parties. Federated learning can play a vital role in this area while maintaining data sovereignty as a large number of participants can create a collaborative ML system incorporating features of all participants' data. 

\subsection{Dataset} 
Wellington camera traps dataset \cite{anton2018monitoring} is used to study the robustness of the FL system. This dataset contains 270,450 images, and 90,150 bursts from 187 camera locations in Wellington, New Zealand. The motion sensor cameras recorded a burst of three images when triggered. The images are taken at night and daytime. The dataset is labelled by citizen scientists or professional ecologists from the Victoria University of Wellington, New Zealand. The images are classified into seventeen categories, with a high degree of imbalance across these categories. There is a varying degree of image quality and about 25\% of the images are empty, which is considered a false positive.  

Several small and camouflaged animals are present in these images, which are very difficult to detect; some example images are shown in Fig.~\ref{fig:ExampleImages}. However, the bursts of images provide an opportunity to detect animals by comparing the images in a single burst. \cite{shashidhara2020sequence}.

\subsection{Experiment Setup}

From the Wellington Camera Trap dataset, we did not consider any image bursts that include less than three images and also did not consider sites that have less than 75 bursts. The image bursts are preprocessed by employing the Mixture of Gaussian (MoG) method and then constructing 3 channel images, as shown in Fig.~\ref{fig:Preprocessing}. We have selected different numbers of sites for training, validation, and testing (discussed in the next section).
For FL experiments, we have used Flower \cite{beutel2022flower} while using the federated averaging method for client model aggregation. Squeezenet CNN \cite{pothos2016fast} is used with Adam optimiser and starting learning rate of 0.0001. Categorical cross-entropy loss function weighted to account for class imbalance and standard augmentations for training batches \cite{shashidhara2020sequence}.  We kept 100 FL iterations with 10 local epochs per iteration and a local batch size of 128.

\begin{figure}[!tb]
    \begin{center}
       {\includegraphics[width=0.9\linewidth]{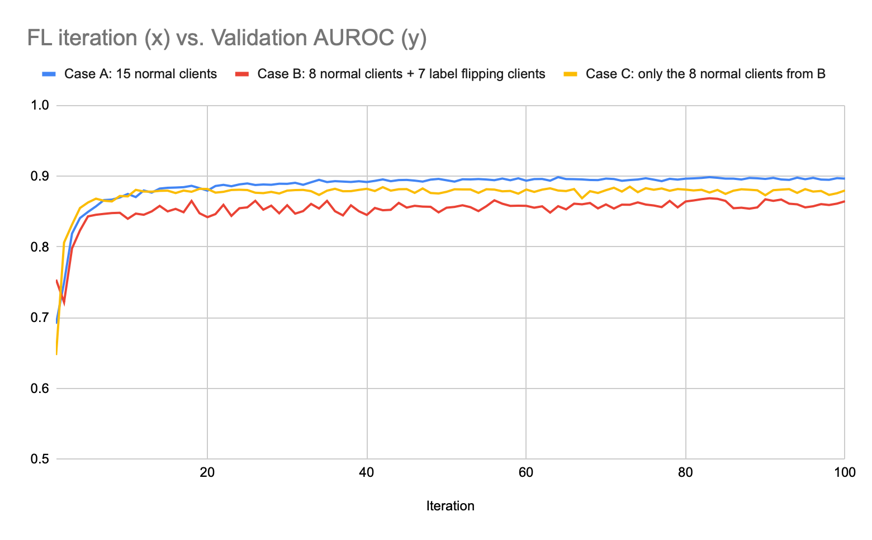}} %\enspace
      \end{center} 
    \caption{FL performance under different numbers of mislabelled training data}
    \label{fig:label_flipping}
\end{figure} 

\begin{figure}[!tb]
    \begin{center}
       {\includegraphics[width=0.95\linewidth]{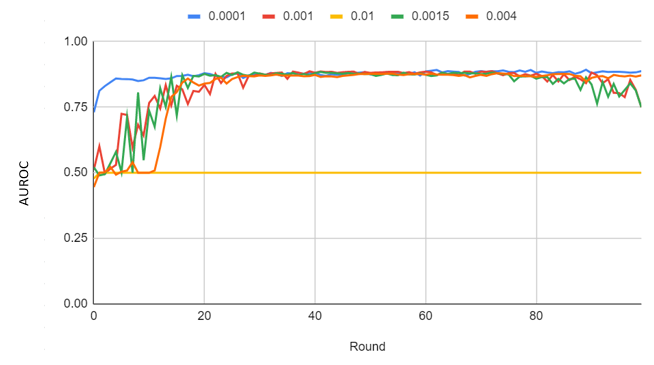}} %\enspace
      \end{center} 
    \caption{FL performance when the learning rate of one client is misconfigured and changed from 0.0001 to 0.01}
    \label{fig:learning_only}
\end{figure}

\begin{figure*}
     \centering
     \begin{subfigure}[b]{0.49\textwidth}
         \centering
         \includegraphics[width=1.0\linewidth]{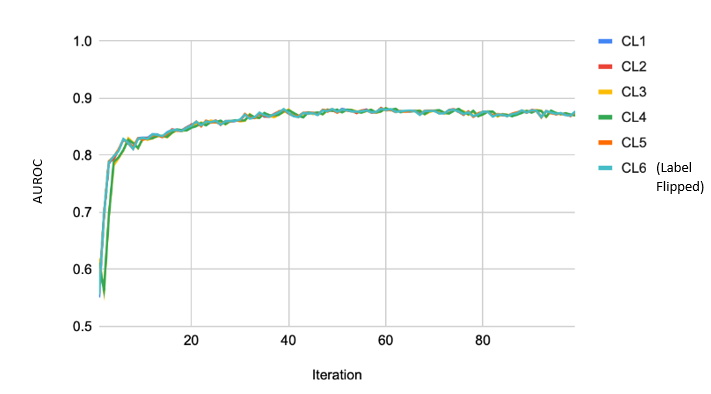}
         \caption{FL performance in the presence of mislabelled data}
         % \caption{All Clients have same learning rate of 0.0001}
         \label{fig:a}
     \end{subfigure}
     \hfill
     \begin{subfigure}[b]{0.49\textwidth}
         \centering
         \includegraphics[width=1.0\linewidth]{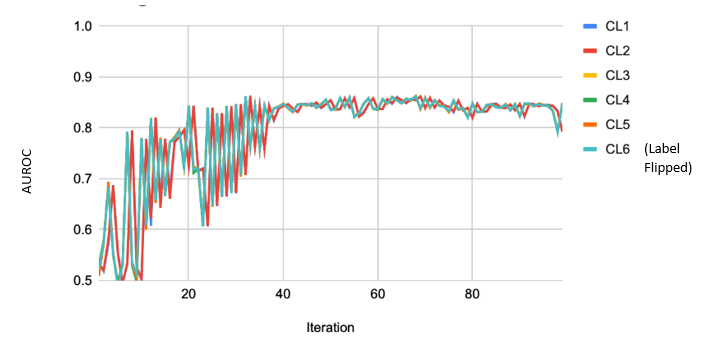}
         \caption{FL performance in the presence of both mislabelled data and misconfigured learning rate of one client (10 times higher)}
         % \caption{Client with mislabelled data has a learning rate 10 times higher than normal clients (0.001)}
         \label{fig:b}
     \end{subfigure}
        \caption{The impact of learning rate on each client in the presence of mislabelled data}
        \label{fig:learning_flipping}
\end{figure*}

\subsection{Results and Discussion}
The following sets of experiments are conducted to observe the robustness of FL system against ML-specific inconsistencies. 
\subsubsection{FL Robustness against Mislabelled data} 
ML-specific inconsistency can be due to low-quality data in terms of mislabelled or flipped label records. Several experiments were conducted to study when one or more clients have mislabelled some or all examples in their training data \cite{jebreel2022defending}. Such mislabelled data can affect the global model when the server aggregates the local parameters trained on them. The inherent inaccessibility of the federated server to the clients and clients' data makes it difficult to detect such issues \cite{lv2022awfc}. 

To evaluate the robustness of FL against such mislabelled clients, we conducted a series of experiments training FL models to compare three scenarios. In the baseline (i.e., Case A), all 15 clients had correctly labelled data. In Case B, 7 clients had mislabelled data while 8 clients had correctly labelled data. Case C involved training the FL model using data only from the 8 clients with correctly labelled data.
% The first set of experiments is conducted to evaluate the robustness of FL against such mislabelled clients. We have compared three cases: Case A as a baseline case with all 15 sites and no mislabelled data, Case B with 7 clients with mislabelled data and 8 clients with correctly labelled data, and Case C where we train the FL model with only 8 clients with correctly labeled data). 

Fig.~\ref{fig:label_flipping} shows the Area Under the Receiver Operating Characteristics (AUROC) values for all FL iterations for all three cases. Case A shows the AUROC values of around 0.9 after the 20th iteration. Case B has around 47\% clients with mislabelled data and shows the AUROC value of 0.86 after the 20th iteration. For case C, where the clients with mislabelled data are removed and the global model considers only 8 normal clients; the AUROC value is around 0.88 after the 20th iteration. We can conclude from the results that the effect of mislabelled data is there on the performance of the FL model but not very significant, which shows the robustness of FL against such attacks. 
Some of these findings do not concur with literature \cite{zhang21}. One factor that is really important to consider here is the learning rate; we kept a low learning rate (0.0001) for all our clients, which keeps the changes to the weights very small due to local learning. We have conducted the second set of experiments (discussed next) to study the effect of learning rate in mislabelled data scenarios. 

\subsubsection{FL Robustness against Misconfiguration}
One ML-specific  inconsistency can be due to the misconfiguration of hyperparameters. One of the hyperparameters that affect the accuracy of the neural networks model is the learning rate, which controls how quickly the model is adapted to the problem \cite{takase2018effective}. Even for FL systems, the learning rate has shown a significant impact on the accuracy of the models and has been used as one of the factors that can be adapted to improve accuracy \cite{xu2021learning}, fight attacks \cite{ozdayi2021defending} and flexible client participation \cite{yang2022anarchic} among other purposes. We have conducted experiments with six normal clients to study the impact of learning rate on FL systems. We changed the learning rate of only one client from 0.0001 to 0.01; the results are shown in Fig.~\ref{fig:learning_only}. The results show that for learning rate up to 0.004, the final AUROC values are not affected and is always around 0.88. However, as the learning rate exceeds 0.004, the maximum AUROC cannot go higher than 0.5. This shows that FL shows high robustness against certain misconfigurations (some clients assigned with a higher local learning rate) to a certain extent but as the values become very high, the system cannot perform well.

\subsubsection{FL Robustness against Mislabelled data and Misconfiguration} 
The next set of experiments is conducted to study the effect of misconfiguration on the performance of the FL model with different learning rates when some clients have mislabelled data. For these experiments, we have chosen five normal clients and one misconfigured client with mislabelled data. The learning rate for the misconfigured client is changed from 0.0001 to 0.001 to observe the effect on the FL system, as shown in Fig.~\ref{fig:learning_flipping}. 

The results show that the higher learning rate plays a significant role in the final AUROC as it reduces from 0.88 to 0.85 when the learning rate of the misconfigured client is changed from 0.0001 to 0.001. Also, in early iterations of FL, the impact is more significant and it takes longer for the system to converge in the second case of a higher learning rate. Hence, an important finding is that low local learning rates impact the robustness of the FL systems. 

\section{Related work}
% related work on unreliable clients 
% As previously discussed, we are studying the effect of unreliable clients on the performance of FL. For this paper, we are considering the unreliability of clients in terms of permanent dropouts, partial dropouts, misconfiguration and low-quality data. 
In this section, we look at some research efforts that tackled unreliable client issues for FL and classify them into two categories: infrastructure-level errors and ML-specific inconsistencies.

\emph{Infrastructure-level errors}: FL requires sharing of local model parameters between clients and a server over multiple learning rounds until the global model converges. In environments with limited and unreliable network resources, such communication may not be possible. As a result, it can affect the training time and further lead to issues such as unstable and/or slow convergence of the global model. 
% A compensation scheme considers similarity among the clients to replace any missing model updates due to unreliable communication with the update from the most similar client where sparse learning is proposed at local clients \cite{mao2022safari}.
A compensation scheme proposed in~\cite{mao2022safari} suggested that any missing model updates can be replaced by the updates from other clients based on their similarity. Their experiments with the CIFAR-10 dataset show high convergence of the global model under varying successful communication probability for ten clients with a learning rate of 0.001 for every client. 
A similar approach has been proposed in~\cite{wang2022friends} where it considered local model updates as ``friends'' when the clients' data distributions are sufficiently similar. To identify ``friends'', the pairwise similarity score among the local model updates is calculated by the FL server. Whenever any client dropout happens, the FL server replaces the missing model update with its ``friends''. Experiments were conducted using MNIST and CIFAR-10 datasets. The results showed that model replacement using the friends' model successfully mitigates the effect of client dropout.
% Clients with different learning rates may pose a challenge when similarity is considered among clients. The clients with sufficiently similar data distribution and hence local model updates are considered ``friends'' for mitigating the impact of client dropouts during the calculation of the global model for FL \cite{wang2022friends}. The pairwise similarity score is calculated among the local model updates of the clients to identify clients that have similar data distribution and are considered friends by the FL server. At any round, when any client dropouts, the FL server replaces the model update of that client with a non-dropping friend. MNIST and CIFAR-10 datasets are used for the experiments, and the results show that friends of the clients are discovered successfully, and the replacement of the models with the friends' model mitigates the effect of client dropout. 
Recently, a secure and efficient FL (SEFL) framework was proposed in~\cite{deng2022secure}. SEFL considers two servers, one for model aggregation and one for managing cryptography primitives. The clients use the weight pruning technique to prune the local model update before sending the encrypted updates to the aggregating server. At the server, the encrypted and pruned updates are homomorphically added. In order to decrypt the aggregated value, the other server is contacted. To ensure the security and privacy of the aggregated model values differential privacy approach is used, which is not discussed any further as it is out of the scope of this work. However, a very interesting and related property of the algorithm is that the aggregation server can train an accurate global model even when only 10\% of the clients share their local updates with the server.      

% FL enables clients to train their local model on local data. Then, the models are aggregated at the server to create a global model with no interaction between the aggregation process and local training. 
\emph{ML-specific inconsistencies}: Data from different clients can be different in terms of quality and label noise. Wrong and/or noisy labels at the clients can negatively affect the global model as the server cannot access local data to filter out the noise. To tackle this problem, different methods have been proposed. For example, a two-level sampling approach from~\cite{wang2022fednoil} allows a) the server to select better client models and b) clients to select clean local data. The confidence score of data samples is calculated using the global model to calculate the overall confidence of the client. The experiments conducted with the CIFAR-10 dataset with 100 clients containing imbalanced data with varying noise ratios show that the proposed approach outperforms several baseline algorithms. FL under Label Noise (FedLN)~\cite{tsouvalas2022federated} estimates per client label noise level and limits the effect of noisy samples on the model's generalisability. At the federated server, Noise-aware Federated Averaging (NA-FedAvg) ~\cite{tsouvalas2022federated} considers estimated clients' noise levels to perform noise-aware aggregation of the client models. Four different datasets are used to test NA-FedAvg with varying numbers of noisy clients with different noise levels. With all noisy clients, the model performance degraded substantially; however, the presence of only 20\% of clean clients enabled the model to perform at 70\% for all noise levels. A different approach for reducing the impact of irrelevant or bad quality clients' data on the performance of FL is a distributed selection method~\cite{tuor2021overcoming} that enables clients to choose only a relevant subset from complete data available at the clients. A model requester is proposed for the FL system that provides a benchmarked dataset to the clients to identify a relevant subset of data for a particular FL task.

\section{Conclusions}
In this paper, we systematically evaluate the impact of unreliable clients on federated learning (FL) performance in rural environments. We investigated different scenarios of unreliable clients including client dropout, misconfiguration, and low-quality training data. Our experiments are conducted on two applications (weeds detection in precision agriculture and wildlife detection in camera traps) using real-world datasets \tl{ and a low number of clients (ranging from 3 to 6). Intuitively, scenarios with a reduced number of clients should be more susceptible to performance degradation in case of one or more clients malfunctioning or misconfiguration. Surprisingly,} experiments show empirical proof that the federated averaging method is resilient to such unreliable clients, both to infrastructure-level and ML-specific inconsistencies. First, the weeds detection case study explored the tolerance to the infrastructure-level category of unreliable clients, such as those with faulty network connections. Second, the animal detection case study (via camera traps) showed the robustness of federated learning towards ML-specific deficiencies such as mislabelled data or incorrect parameter values. We have shown that despite its simplicity, federated averaging learning should be the algorithm of choice for distributed training environments with unreliable rural infrastructure and non-technical users. 

Our next step involves investigating the applicability of techniques such as Reinforcement Learning as part of the client selection to further improve the performance of Federated Averaging by selecting the best data sources/clients to train the global model. We will further develop platforms and approaches to support the real-world deployment of FL. The platform we used in this paper using Function-as-a-Service provides the first step towards this goal. By introducing higher level systems, programming abstractions, and libraries, we will enable the optimisation workloads and resources of FL in real-world systems.

\bibliographystyle{IEEEtran}
\bibliography{CLOUD.bib}

\end{document}